\documentclass[12pt]{article}
\usepackage{amsmath,amsthm,amssymb,euscript,epsf,epsfig,color,graphicx}
\usepackage{array}
\usepackage{fancybox}
\newcommand{\ov}{\overrightarrow}

\newcommand{\be}{\begin{equation}}
\newcommand{\ee}{\end{equation}}
\newcommand{\bea}{\begin{eqnarray}\displaystyle}
\newcommand{\eea}{\end{eqnarray}}

\makeatletter
\@addtoreset{equation}{section}
\makeatother
\renewcommand{\theequation}{\thesection.\arabic{equation}}
\def\one{{\hbox{ 1\kern-.8mm l}}}
\def\zero{{\hbox{ 0\kern-1.5mm 0}}}

  \def\cL{{\cal L}}

 \def\cZ{{\cal Z}}

\begin{document}

\makeatletter
\@addtoreset{equation}{section}
\makeatother
\renewcommand{\theequation}{\thesection.\arabic{equation}}

\rightline{QMUL-PH-17-03}
\vspace{1.8truecm}

\vspace{10pt}


{\LARGE{ 
\centerline{\bf  Linguistic Matrix Theory } 
}}  

\vskip.5cm 

\thispagestyle{empty} \centerline{
    {\large \bf Dimitrios Kartsaklis${}^{a,}$\footnote{ {\tt d.kartsaklis@qmul.ac.uk}}, }
   {\large \bf  Sanjaye Ramgoolam${}^{b,c,}$\footnote{ {\tt s.ramgoolam@qmul.ac.uk}},    }
               {\large \bf  Mehrnoosh Sadrzadeh${}^{a,}$\footnote{ {\tt mehrnoosh.sadrzadeh@qmul.ac.uk }}    }
                                                       }

\vspace{.4cm}
\centerline{{\it ${}^a$ School of Electronic Engineering and Computer Science, }}
\centerline{{\it Queen Mary University of London,  }}
\centerline{{\it  Mile End Road, } }
\centerline{{\it London E1 4NS, UK } }

\vspace{.2cm}
\centerline{{\it ${}^b$ Centre for Research in String Theory, School of Physics and Astronomy},}
\centerline{{ \it Queen Mary University of London},} \centerline{{\it
    Mile End Road, London E1 4NS, UK}}
    
    \vspace{.2cm}
\centerline{{\it ${}^c$ National Institute for Theoretical Physics,}}
\centerline{{\it School of Physics and Centre for Theoretical Physics, }}
\centerline{{\it University of the Witwatersrand, Wits, 2050, South Africa } }

\vspace{.5truecm}

\thispagestyle{empty}

\centerline{\bf ABSTRACT}

\vskip.2cm 

Recent research in computational linguistics has developed algorithms which associate matrices with adjectives and verbs, based on 
the distribution of words in a corpus of text. These matrices are linear operators on a vector space of context words. They are used to  construct the meaning of composite expressions from that of the elementary constituents, forming part of a compositional distributional approach to semantics. 
 We propose a Matrix Theory approach to this data, based on permutation symmetry along with  Gaussian weights and their perturbations. A simple Gaussian model is  tested against word matrices created from a large corpus of text. We characterize the cubic and quartic departures from the model, which we propose, alongside the  Gaussian parameters, as signatures for comparison of linguistic corpora. We propose that perturbed Gaussian models with permutation symmetry provide a promising framework for characterizing the nature of universality in the statistical properties of word matrices.  
 The  matrix theory framework developed here  exploits the view of  statistics as zero dimensional perturbative quantum field theory. It perceives language as a physical system realizing a  universality class of matrix statistics characterized by permutation symmetry. 

\setcounter{page}{0}
\setcounter{tocdepth}{2}

\newpage

\tableofcontents

\section{Introduction}

Meaning representation is a task at the core of Computational Linguistics research. At the word level, models based on the so-called {\em distributional hypothesis} (the meaning of a word is determined by the contexts in which it occurs) \cite{Harris,Firth} associate meaning with vectors of statistics reflecting the co-occurrence of the word with a set of contexts. While distributional models of this form have been proved very useful in evaluating the semantic similarity of words by application of vector algebra tools \cite{Rubenstein,Salton}, their statistical nature do not allow them to scale up to the level of multi-word phrases or sentences. Recent methods \cite{GSC1003,GS1106,Maillard,Baroni,KartSadrPul2013} address this problem by adopting a compositional approach: the meaning of relational words such as verbs and matrices is associated with matrices or higher order tensors, and composition with the noun vectors takes the form of tensor contraction. In tensor-based models of this form, the grammatical type of each word determines the vector space in which the word lives: take $N$ to be the noun space and $S$ the sentence space, then an adjective becomes a linear map $N\to N$ living in $N^* \otimes N$, an intransitive verb a map $N \to S$ in $N^* \otimes S$, and a transitive verb a tensor of order 3 in $N^* \otimes S \otimes N^* $. Hence, given a transitive sentence of the form ``John likes Mary'', vectors $\ov{John}$, $\ov{Mary}$ representing the meaning of the noun arguments and an order-3 tensor $M_\text{likes}$ for the verb, the meaning of the sentence is a vector in $S$ computed as $\ov{John}M_\text{likes}\ov{Mary}$. 

Given this form of meaning representation, a natural question is how to characterize the distribution of the matrix entries for all the relational words in the corpus, which correspond to a vast amount of data. Our approach to this problem is informed by Random Matrix theory. Random matrix theory  has a venerable history starting from Wigner and Dyson \cite{Wigner,Dyson}  who used it to describe the distribution of energy levels of complex nuclei. A variety of physical data in diverse physical systems has been shown to obey random matrix statistics. The matrix models typically considered have continuous symmetry groups which relate the averages and dispersions of diagonal and off-diagonal elements of the matrix elements. Our study of these averages in the context of language shows that there are significant differences between these characteristics for diagonal and off-diagonal elements. 

This observation motivates the study of a simple class of solvable Gaussian models without continuous symmetry groups. In the vector/tensor space models of language meaning, it is natural to expect a discrete symmetry of permutations of the context words used to define the various vectors and tensors. Random matrix integrals also arise in a variety of applications in theoretical physics, typically as the reductions to zero dimensions from path integrals of a higher dimensional quantum field theory. 
We develop a matrix theory approach to linguistic data, which draws on random matrix theory as well as quantum field theory, and where the permutation symmetry plays a central role. 

\vspace{0.2cm}
The paper is organised as follows:

Section 2 gives some more detailed background on how random matrices arise in applied and theoretical physics, 
highlighting the role of invariant functions of matrix variables in the definition of the probability measure
 and  the observables. 

Section 3 describes the main ideas behind distributional models of meaning at the word level, and explains how the principle of compositionality can be used to lift this concept to the level of phrases and sentences.

Section 4 discusses in detail the setting used in the experimental work of this paper, explaining the process applied for creating the dataset and providing technical details about constructing vectors for nouns and matrices for verbs and adjectives by application of linear regression.
 
Section 5 presents data on distributions of several selected matrix elements which motivates us to consider Gaussian measures as a starting point for connecting $S_D$ invariant probability distributions with the data. 
Section 6 describes a 5-parameter Gaussian model. Section 7 discusses the comparison of the  theory with data. 
Finally, Section 8 discusses future directions. 

\section{Random matrices: Observables and symmetries} 

The association of relational words such as adjectives and verbs in a corpus with matrices produces a large amount of matrix data, and raises the question of characterising the information present in this data. Matrix distributions have been studied in a variety of areas of applied and 
theoretical  physics. Wigner and Dyson studied the energy levels of complex nuclei, which are eigenvalues of hermitian matrices. The techniques 
they developed have been applied to complex atoms, molecules, subsequently to scattering matrices, chaotic systems amd financial correlations. 
Some references which will give an overview of the theory and diversity of applications of random matrix theory are \cite{mehta,GMW97,Beenakker,EY}. 
The spectral studies of Wigner and Dyson focused on systems with continuous symmetries, described by unitary, orthogonal or symplectic groups.

Matrix theory has also seen a flurry of applications in fundamental physics, an important impetus coming from the AdS/CFT correspondence \cite{Malda}, which gives an  equivalence between four dimensional quantum field theories and  ten dimensional string theory. These QFTs have conformal invariance and quantum states correspond to polynomials in matrix fields $ M ( \vec x , t ) $,  invariant under gauge symmetries, such as the unitary groups. Thanks to conformal invariance, important observables in the string theory are related to quantities which can be computed in reduced matrix models  where the quantum field theory path integrals simplify to ordinary matrix integrals (for  reviews of these directions in AdS/CFT see \cite{Agmoo,PermGIs}).  This sets us back to  the world of matrix distributions. These matrix integrals also featured in earlier versions of gauge-string duality for low-dimensional strings, where they find applications in the topology of moduli spaces of Riemann surfaces (see \cite{GinspargMoore} for a review). 

An archetypal  object of study in these areas of applied and theoretical physics is the matrix integral 
\bea\label{basicGM}
\cZ ( M ) = \int dM e^{ - tr M^2 }  
\eea
which defines a Gaussian Matrix distribution, and the associated  matrix moments 
\bea\label{simplestGaussMat} 
\int dM e^{ - tr M^2 } tr M^k 
\eea
Moments generalizing the above are relevant to graviton interactions in ten-dimensional string theory in the context of AdS/CFT. 
Of relevance to the study of spectral data, these moments contain information equivalent to eigenvalue distributions, since the 
matrix measure can be transformed to a measure over eigenvalues using an appropriate Jacobian for the change of variables. 
 More generally, perturbations of 
the Gaussian matrix measure are of interest:
\bea\label{perturbedGM}
\cZ ( M , g ) = \int dM e^{ - tr M^2 + \sum_{ k } g_k tr M^k } 
\eea
for coupling constants $g_k$. In the higher dimensional quantum field theories, the $g_k$ are coupling constants controlling the interaction 
strengths of particles. 

In the linguistic matrix theory we develop  here, we  study the matrices coming from linguistic data using Gaussian distributions generalizing (\ref{basicGM})-(\ref{perturbedGM}). The matrices we use are not hermitian or real symmetric; they are general real matrices. Hence, a distribution of 
eigenvalues is not the natural  way to study their statistics. Another important property of the application at hand is that while it is natural to 
consider matrices of a fixed size $ D \times D$, there is no reason to expect the linguistic or statistical properties of these matrices to 
be invariant under a continuous symmetry. It is true that dot products of vectors  (which are used 
in measuring word similarity in distributional semantics) 
are invariant under the continuous  orthogonal group $O(D)$. However,  in general we may expect no more than an invariance under the discrete  symmetric group $S_D$  of $D!$ permutations of the basis vectors.

The general framework for our investigations will therefore be 
Gaussian matrix integrals of the form:
\bea 
\int dM e^{  L ( M ) + Q ( M ) + \textit{perturbations } }   
\eea
where  $L(M)$ is a linear function of the matrix $M$, invariant under $S_D$, and $Q(M)$ is a quadratic function invariant under $ S_D$. Allowing linear terms in the Gaussian action means that the matrix model can accommodate data which has non-zero expectation value. The quadratic terms are eleven in number for $ D \ge 4$, but we will focus on a simple solvable subspace which involves three of these quadratic invariants 
 along with two linear ones.  Some neat $S_D$ representation theory behind  the enumeration of these invariants is explained in Appendix \ref{sec:SDMatrixInvts}.  
 The 5-parameter model is described in Section \ref{sec:5parameters}. 

\section{Vectors and tensors in linguistics: Theory}\label{Sec:Expt} 

\subsection{Distributional models of meaning}

One of the most successful models for representing meaning at the word level in computational linguistics is {\em distributional semantics}, based on the hypothesis that the meaning of a word is defined by the contexts in which it occurs \cite{Harris,Firth}. In a distributional model of meaning, a word is represented as a vector of co-occurrence statistics with a selected set of possible contexts, usually single words that occur in the same sentence with the target word or within a certain distance from it. The statistical information is extracted from a large corpus of text, such as the web. Models of this form have been proved quite successful in the past for evaluating the semantic similarity of two words by measuring (for example) the cosine distance between their vectors. 

For a word $w$ and a set of contexts $\{ c_1, c_2, \cdots, c_n \}$, we define the distributional vector of $w$ as:

\begin{equation}
   \ov{w} = (f(c_1),f(c_2),\cdots, f(c_n))
\end{equation}

\noindent where in the simplest case $f(c_i)$ is a number showing how many times $w$ occurs in close proximity to $c_i$.  In practice, the raw counts are usually smoothed by the application of an appropriate function such as {\em point-wise mutual information} (PMI). For a context word $c$ and a target word $t$, PMI is defined as:

\begin{equation}
   PMI(c,t) = \log \frac{p(c,t)}{p(c)p(t)} = \log \frac{p(c|t)}{p(c)} = \log \frac{p(t|c)}{p(t)} = \log \frac{count(c,t) \cdot N}{count(t)\cdot count(c)}
   \label{equ:PMI}
\end{equation}

\noindent where $N$ is the total number of tokens in the text corpus, and $count(c,t)$ is the number of $c$ and $t$ occurring in the same context. 
 The intuition behind PMI is that it provides a measure of how often two events (in our case, two words) occur together, with regard to how often they occur independently. Note that a negative PMI value implies that the two words co-occur {\em less often} that it is expected by chance; in practice, such indications have been found to be less reliable, and for this reason it is more common practice to use the positive version of PMI (often abbreviated as PPMI), in which all negative numbers are replaced by 0.

One problem with distributional semantics is that, being purely statistical in nature, it does not scale up to larger text constituents such as phrases and sentences: there is simply not enough data for this. The next section explains how the {\em principle of compositionality} can be used to address this problem. 

\subsection{Grammar and tensor-based models of meaning}

The starting point of tensor-based models of language is  a formal analysis of the grammatical structure of  phrases and sentences. These  are then combined with a  semantic analysis, which assigns meaning representations to words and extends them to phrases and sentences  compositionally, based on the grammatical analysis:
\[
\mbox{Grammatical Structure} {\implies} \mbox{Semantic Representation}
\]

The grammatical  structure of language  has been made  formal  in  different ways by different linguists. We have the work of Chomsky on generative grammars \cite{Chomsky}, the original functional systems of Ajdukiewicz \cite{Ajdukiewicz} and Bar-Hillel \cite{Bar-Hillel} and their type logical reformulation by Lambek in  \cite{Lambek}. There is a variety of  systems that build on the work of Lambek, among which the Combinatorial Categorial Grammar (CCG) of Steedman \cite{Steedman} is the most widespread. These latter models employ   ordered algebras, such as residuated monoids, elements of which are interpreted as function-argument structures. As we shall see in more detail below, they   lend themselves well  to a semantic theory in terms of  vector and tensor  algebras via the map-state duality:
\[
\mbox{Ordered Structures} {\implies} \mbox{Vector and Tensor Spaces}
\]

There are  a few choices around  when it comes to which formal grammar to use as base. We discuss two possibilities:  pregroup grammars and CCG. 
The contrast between the two lies in the fact that pregroup grammars have  an underlying ordered structure and form a partial order compact closed category \cite{KellyLaplaza}, which enables us to formalise  the  syntax-semantics passage  as a strongly monoidal functor, since the category of finite dimensional vector spaces and linear maps is also compact closed (for details see \cite{GSC1003, CoeckeGrefenSadr,PrellerLambek}). So we have:

\[
\mbox{Pregroup Algebras} \stackrel{\mbox{\small strongly monoidal functor}}{\implies} \mbox{Vector and Tensor Spaces}
\]

In contrast, CCG  is based on a set of rules,  motivated by the combinatorial calculus of Curry \cite{Curry} that was developed for reasoning about functions in arithmetics and  extended and altered for purposes of reasoning about natural language constructions.   The CCG is  more expressive than pregroup grammars: it covers the weakly context sensitive fragment of  language  \cite{Weir}, whereas pregroup grammars  cover the context free fragment \cite{Busz}. 


\subsection{Pregroup grammars}

A pregroup algebra  $(P, \leq, \cdot, 1, (-)^r, (-)^l)$ is a partially ordered monoid where each element $p \in P$ has a left $p^l$ and right $p^r$ adjoint, satisfying the following  inequalities:
\[
p \cdot p^r \leq 1 \leq p^r \cdot p \qquad p^l \cdot p \leq 1 \leq p \cdot p^l
\]
A pregroup grammar over a set of words $\Sigma$  and a set of basic grammatical types is denoted by $({\mathcal P}({\mathcal B}), {\mathcal R})_{\Sigma}$ where ${\mathcal P}({\mathcal B})$ is a pregroup algebra generated over  $\mathcal B$ and  $ {\mathcal R}$ is a relation $\mathcal R \subseteq \Sigma \times P({\mathcal B})$  assigning to each word a set of grammatical types. This relation is otherwise known as a \emph{lexicon}. Pregroup grammars were  introduced by Lambek \cite{Lambek1}, as a simplification of his original Syntactic Calculus \cite{Lambek}.

As an example,  consider the set of words $\Sigma = \{$men, cats, snore, love, sees, tall$\}$ and the set of basic types ${\mathcal B} = \{n, s\}$ for $n$ a  noun phrase and $s$ a declarative sentence. The pregroup grammar over $\Sigma$ and ${\mathcal B}$ has the following lexicon:
\[
\left\{(\textmd{men}, n), (\textmd{cats}, n), (\textmd{tall}, n\cdot n^l), 
 (\textmd{snore}, n^r \cdot s),
 (\textmd{love}, n^r \cdot s \cdot n^l)\right\}
\]

Given a string of words $w_1, w_2, \cdots, w_n$, its pregroup grammar derivation is the following inequality, for $(w_i, t_i)$  an element of the lexicon of $P({\mathcal B})$ and $t \in P({\mathcal B})$:
\[
t_1 \cdot t_2 \cdot \cdots t_n \leq t
\]
When the string is a sentence, then $t = s$.  For example, the  pregroup derivations for the the phrase ``tall men'' and  sentences ``cats snore'' and ``men love cats'' are as follows: 
\begin{align*}
&(n \cdot n^l) \cdot n \leq n \cdot 1 = n\\
&n \cdot (n^r \cdot s) \leq 1 \cdot s = s\\
&n \cdot (n^r \cdot s \cdot n^l) \cdot n \leq 1 \cdot s \cdot 1 =  s
\end{align*}
In this setting the partial order is read as grammatical reduction. For instance, the juxtaposition of types of the words of a grammatically formed sentence reduces to the type $s$, and a grammatically formed noun 
phrase reduces to the type $n$.  

%
%

\subsection{Combinatorial Categorial Grammar}

Combinatorial Categorial Grammar has a set of  atomic and complex  categories and a set of rules. Complex types are formed from atomic types  by using  two slash operators $\setminus$ and $/$; these are employed  for function construction and implicitly encode  the grammatical order of words in phrases and sentences.  

Examples of atomic categories are the  types of noun phrases $n$ and  sentences $s$. Examples of complex categories are the types of adjectives and intransitive verbs, $n/n$  and  $s\setminus n$, and the type of transitive verbs, $(s\setminus n)/n$.  The idea behind these assignments is that a word with a complex type $X \setminus Y$ or $X / Y$ is a function that takes an argument of the form $Y$ and returns a result of type $X$.  For example,  adjectives and intransitives  verb are  encoded as unary functions: an adjective such as ``tall'' takes a noun phrase of type $n$, such as ``men'' on its right  and return a  modified noun phrase $n$, i.e.  ``tall men''. On the other hand, an intransitive verb such as ``snore'' takes a noun  phrase  such as ``cats'' on its left and returns the sentence ``cats snore''. A transitive verb, such as ``loves'' first takes an argument of type $n$ on its right, e.g. ``cat'', and produces a function of type $s\setminus n$, for instance ``love cats'', then takes an argument of type $n$ on its left, e.g. ``men'' and produces a sentence, e.g. ``men love cats''. 

In order to combine words  and form phrases and sentences, CCG  employs a set of rules.  The two  rules which formalise  the reasoning in the above examples are called \emph{applications} and are as follows:
\[
(>)\ X / Y  \quad Y \implies X \hspace {2cm}
(<)\ Y \quad X \setminus Y \implies X
\]

Using this rule, the  above examples  are formalised as follows:

\begin{center}
\vspace{0.2cm}
\begin{tabular}{ccccc}
&tall & men&& \\
&$n/n$ & $n$ & $\stackrel{>}{\implies}$ & $n$\\
&cats & snore&&\\
&$n$ & $s\setminus n$ & $\stackrel{<}{\implies}$ & $s$\\
 men & love & cats&&\\
$n$ & $(s\setminus n)/n$ & $n$ & $\stackrel{>}{\implies}$& $n \quad s\setminus n \stackrel{<}{\implies} n$
\end{tabular}
\end{center}

\subsection{Semantics}
\label{sec:semantics}

On the semantic side, we  present material  for the CCG types. There is a translation $t$ between the CCG and pregroup types, recursively defined as follows:
\[
t(X /Y) :=  t(X) \cdot t(Y)^l \qquad
t(X\setminus Y) :=   t(Y)^r \cdot t(X)
\]
This will enable us to use this semantics for pregroups as well. For a detailed semantic development based on pregroups, see \cite{GSC1003,CoeckeGrefenSadr}. 

We  assign vector and tensor spaces to each type and assign the elements of these spaces  to the words of that type.  Atomic types are assigned atomic vector spaces (tensor order 1), complex types are assigned tensor spaces with rank equal to the number of slashes of the type. 

Concretely,  to an atomic type,   we assign  a  finite atomic  vector space $U$; to a complex type $X/Y$ or $X\setminus Y$, we assign  the tensor space    $U \otimes U$. This is a formalisation of the fact that in vector spaces   linear maps are,  by the map-state duality,  in correspondence with elements of  tensor spaces.  So, a noun such as ``men'' will get assigned a vector in the space $U$, whereas an adjective such as ``tall'' gets assigned a matrix, that is  a linear map from $U$ to $U$. Using the fact that in finite dimensional spaces 
choosing an orthonormal basis identifies  $U^* $ and $ U$, this map  is an element of $U \otimes U$.   Similarly, to an intransitive verb ``snores'' we assign a matrix:  an element  of the tensor space $U \otimes U$, and to a transitive verb ``loves'', we assign a ``cube'': an element of the tensor space  $U \otimes U \otimes U$.\footnote{Note that, for simplicity, in this discussion we do not differentiate between the noun space and the sentence space; this is a common practice in compositional models, see for example \cite{KartSadrPul2013}.}
 
The rules of CCG are encoded as  tensor contraction. In general, for $T_{ij \cdots z}$ an element of $X\otimes Y \otimes \cdots \otimes Z$ and $V_j$ an element of $Z$, we have that:
\[
T_{ij \cdots z} V_z
\]
In particular, given an adjective $Adj_{ij} \in U \otimes U$ and a noun $V_j \in U$, the corresponding   noun  vector  $Adj_{ij} V_j$, e.g. ``tall men'' is  $tall_{ij} men_j$.  Similarly, an intransitive verb $Itv_{ij} \in U \otimes U$ is applied to a  noun $V_j \in U$ and forms the sentence $Itv_{ij}  V_j$, e.g. for  ``mens snore'' we obtain the vector $snore_{ij}men_j$. Finally, a transitive verb $Tv_{ijk} \in U \otimes U \otimes U$ applies to  nouns $W_j, V_k \in U$ and forms a transitive sentence $(Vrb_{ijk} V_k) W_j$, e.g. ``cats chase mice'' corresponds the vector $(chase_{ijk} mice_k)cats_j$.

Given this theory,  one needs to concretely implement  the vector space $U$ and build vectors and tensors for words. The literature contains a variety of methods, ranging from  analytic ones that combine arguments of  tensors \cite{GS1106,KartSadrPul2013,KartSadrBalk2016,calco2015},  linear  and multi linear regression methods  \cite{Baroni,BarZam2010,GrefenSadrBar2011},  and  neural networks \cite{Maillard}. The former methods are quite specific; they employ assumptions about their underlying spaces that  makes them unsuitable for the general  setting of random matrix theory. The neural networks methods used for building tensors are a relatively new development and not widely tested in practice.  In the next section, we will thus go through the linear regression method described in \cite{Baroni, BarZam2010}, which is considered a standard approach for similar tasks.

%
%

\section{Vectors and   tensors in  linguistics: Practice}\label{sec:Expt} 

Creating vectors and tensors representing the meaning of words requires a text corpus sufficiently large to provide reliable statistical knowledge. For this work, we use a concatenation of the ukWaC corpus (an aggregation of texts from web pages extracted from the .uk domain) and a 2009 dump of the English Wikipedia\footnote{http://wacky.sslmit.unibo.it/doku.php?id=corpora}---a total of 2.8 billion tokens (140 million sentences). The following sections detail how this resource has been used for the purposes of the experimental work presented in this paper.

\subsection{Preparing a dataset}
\label{sec:dataset}

We work on the two most common classes of content words with relational nature: verbs and adjectives. Our study is based on a representative subset of these classes extracted from the training corpus, since the process of creating matrices and tensors is computationally expensive and impractical to be applied on the totality of these words. The process of creating the dataset is detailed below:

\begin{enumerate}
  \item We initially select all adjectives/verbs that occur at least 1000 times in the corpus, sort them by frequency, and discard the top 100 entries (since these are {\em too} frequent, occurring in almost every context, so less useful for the purpose of this study). This produces a list of 6503 adjectives and 4385 verbs.
  \item\label{step:sel} For each one of these words, we create a list of arguments: these are nouns modified by the adjectives in the corpus, and nouns occurring as objects for the verb case. Any argument that occurs less than 100 times with the specific adjective/verb is discarded as non-representative.
  \item We keep only adjectives and verbs that have at least 100 arguments according to the selection process of Step \ref{step:sel}. This produced a set of 273 adjectives and 171 verbs, which we use for the statistical analysis of this work. The dataset is given in Appendix \ref{app:dataset}.
\end{enumerate}

The process is designed to put emphasis on selecting relational words (verbs/adjectives) with a sufficient number of relatively frequent noun arguments in the corpus, since this is very important for creating reliable matrices representing their meaning, a process described in Section \ref{sec:lr}.

\subsection{Creating vectors for nouns}
\label{sec:vecs}

The first step is the creation of distributional vectors for the nouns in the text corpus, which grammatically correspond to atomic entities of language, and will later form the raw material for producing the matrices of words with relational nature, i.e. verbs and adjectives. The basis of the noun vectors consists of the 2000 most frequent {\em content} words in the corpus,\footnote{Or subsets of the 2000 most frequent content words for lower dimensionalities.} that is, nouns, verbs, adjectives, and adverbs. The elements of the vector for a word $w$ reflect co-occurrence counts of $w$ with each one the basis words, collected from the immediate context of $w$ (a 5-word window from either side of $w$), for each occurrence of $w$ in the training corpus. As it is common practice in distributional semantics, the raw counts have been smoothed by applying positive PMI (Equation \ref{equ:PMI}). Based on this method, we create vectors for all nouns occurring as arguments of verbs/adjectives in our dataset.

\subsection{Creating matrices for verb and adjectives}
\label{sec:lr}

Our goal is to use the noun vectors described in Section \ref{sec:vecs} in order to create appropriate matrices representing the meaning of the verbs and adjectives in a {\em compositional} setting.  For example, given an adjective-noun compound such as ``red car'', our goal is to produce a matrix $M_{red}$ such that $M_{red}\ov{car} = \ov{y}$, where $\ov{car}$ is the distributional vector of ``car'' and $\ov{y}$ a vector reflecting the distributional behaviour of the compound ``red car''. Note that a non-compositional solution for creating such a vector $\ov{y}$ would be to treat the compound ``red car'' as a single word and apply the same process we used for creating the vectors of nouns above \cite{Baroni, BarZam2010}. This would allow us to create a dataset of the form $\{ (\ov{car},\ov{red~car}), (\ov{door},\ov{red~door}), \cdots \}$ based on all the argument nouns of the specific adjective (or verb for that matter); the problem of finding a matrix which, when contracted by the vector of a noun, will approximate the distributional vector of the whole compound, can be solved by applying multi-linear regression on this dataset.\footnote{Note that this non-compositional method cannot be generalized for text segments longer than 2-3 words, since data sparsity problem would prevent us for creating reliable distributional vectors for the compounds.}

Take matrices $X$ and $Y$, where the rows of $X$ correspond to vectors of the nouns that occur as arguments of the adjective, and the rows of $Y$ to the distributional vectors of the corresponding adjective-noun compounds. We would like to find a matrix $M$ that minimizes the distance of the predicted vectors from the actual vectors (the so-called {\em least-squares error}), expressed in the following quantity:

\begin{equation}
  \frac{1}{2m}\left( ||MX^T-Y^T||^2+\lambda ||M||^2 \right)
  \label{equ:lr}
\end{equation}

\noindent 
where $m$ is the number of arguments, and $\lambda$ a regularization parameter that helps in avoiding {\em overfitting}: the phenomenon in which the model memorizes perfectly the training data, but performs poorly on unseen cases. This is an optimization problem that can be solved by applying an iterative process such as {\em gradient descent}, or even analytically, by computing $M$ as below:

\begin{equation}
 M = (X^T X)^{-1} X^T Y
\end{equation}

In this work, we use gradient descent in order to produce matrices for all verbs and adjectives in our dataset, based on their argument nouns. For each word we create $D\times D$ matrices for various $D$s, ranging from 300 to 2000 dimensions in steps of 100; the different dimensionalities will be used later in Section \ref{sec:comparison} which deals with the data analysis. The selection procedure described in Section \ref{sec:dataset} guarantees that the argument set for each verb/adjective will be of sufficient quantity and quality to result in a reliable matrix representation for the target word. 

It is worth noting that the elements of a verb or adjective matrix created with the linear regression method do not directly correspond to some form of co-occurrence statistics related to the specific word; the matrix acts as a linear map transforming the input noun vector to a distributional vector for the compound. Hence, the ``meaning'' of verbs and adjectives in this case is not directly distributional, but {\em transformational}, along the premises of the theory presented in Section \ref{sec:semantics}.

\section{Permutation symmetric Gaussian matrix models} 

Gaussian matrix models rely on the assumption that matrix elements follow Gaussian distributions. In the simplest models, such as the one in (\ref{simplestGaussMat}), we have equal means and dispersions for the diagonal and off-diagonal matrix elements. In the past, analytic calculations have been applied to obtain eigenvalue distributions, which were compared with data. In this work we take a different approach: a blend between statistics and effective quantum field theory guided by symmetries. 

\begin{figure}[t!]
  \includegraphics[trim={2cm 1.8cm 2cm 1.8cm},clip,width=\linewidth]{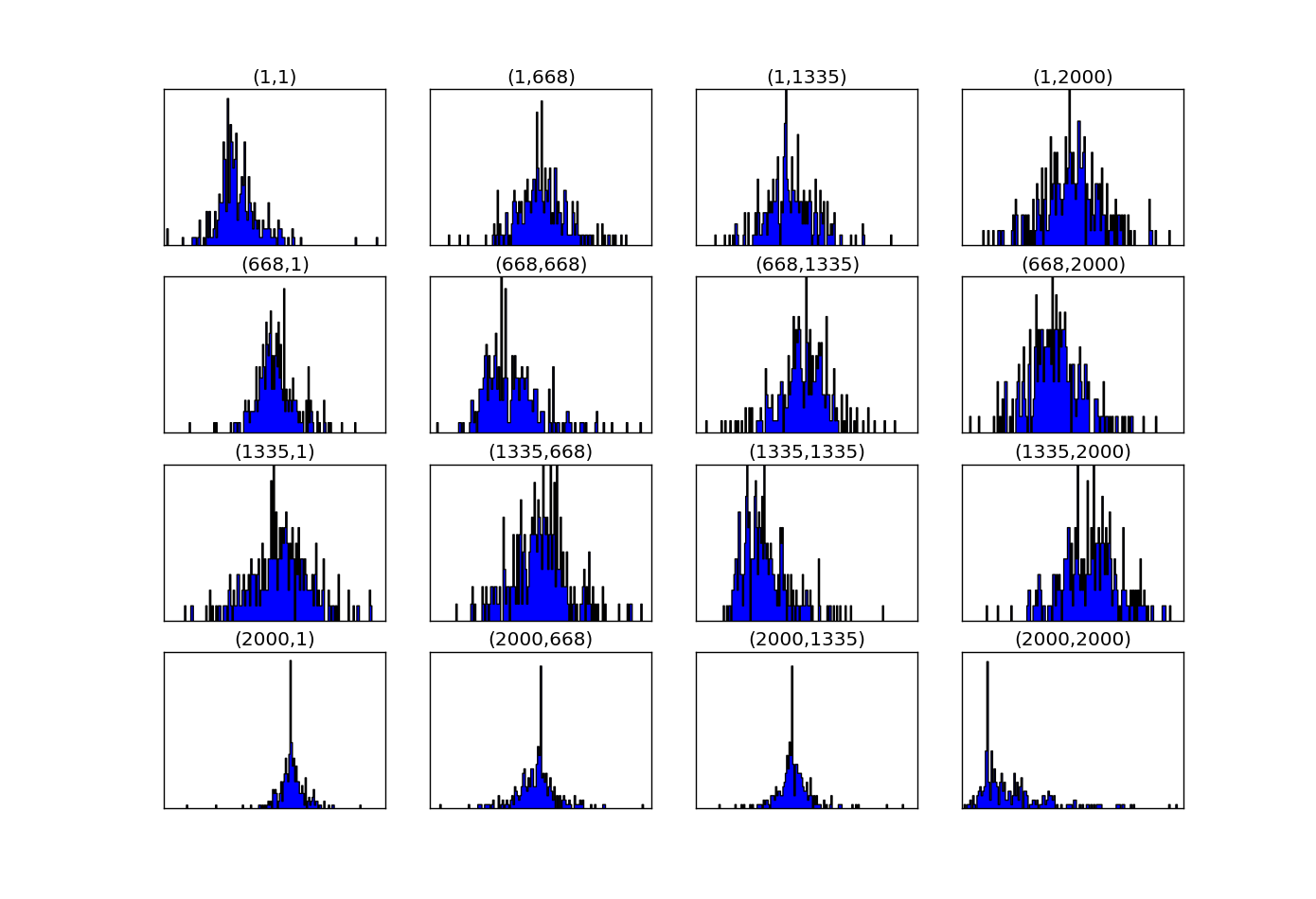}
  \caption{Histograms for elements of adjective matrices}
  \label{fig:GME}
\end{figure}

We start by a qualitative evaluation of the distribution of elements from adjective matrices of size $2000\times 2000$, created as detailed in Section \ref{sec:lr}. In Figure \ref{fig:GME} we plot histograms for different $M_{ij}$'s corresponding to selected $(i,j)$ pairs. Each histogram shows the distribution of the specific element across all adjectives in our dataset; each bar along the horizontal axis corresponds to a different value interval, while the height of the bar reflects frequency (i.e. for how many words the specific element falls within the interval represented by the bar). These histograms, presented here for demonstrative purposes,  look qualitatively like Gaussians. 

This motivates an investigation of Gaussian statistics for word matrices, along the lines of random matrix theory.  
 In the simplest physical applications, the means and variances of diagonal and off diagonal elements are equal for 
 reasons related to underlying continuous symmetries. 
  When $M$ is a hermitian operator corresponding to the Hamiltonian of a quantum system described by states in a Hilbert space, there is a unitary symmetry preserving the inner product. Invariants under unitary group symmetries in this case are traces. The quadratic invariant $tr M^2$ sums diagonal and off-diagonal elements with equal weights. In the linguistic context, there is no reason to expect continuous symmetry groups constraining the statistics. On the other hand,  when the matrices for adjectives or intransitive verbs are constructed by using frequencies of co-occurrences and linear regression, there is no particular significance to the order of context words which correspond to 
 the basis elements of the vector space, where the word matrix $M$ is a linear operator. It is therefore natural to consider a smaller symmetry:  the permutations in the symmetric group $S_D$, a finite symmetry group as opposed to the $D^2$ dimensional manifold of group elements in the unitary group $U(D)$. One can even imagine breaking the symmetry further by choosing  the context words to be ordered according to their own frequency in the corpus. We do not make such special choices in our experiments. This gives very good motivation to consider the universality class of  permutation symmetric models, based on the application to distributional models of meaning we are developing. As far as we are aware, such models have not been systematically studied in physics applications.  Of course the framework of $S_D$ invariant models includes the more restricted models with larger symmetry at special values of the parameters. 
 
The initial inspection of histograms for individual matrix elements provides a strong argument in favour of considering Gaussian matrix models in the linguistic context. A theoretical argument can be sought from 
the central limit theorem, which gives general mechanisms for 
Gaussian random variables to arise as the sums of other variables with finite mean and norm. A possible objection to this is 
that slowly decaying power laws with infinite variance are typical in linguistics, with Zipf's law\footnote{Zipf's law \cite{Zipf} is empirical and states that the frequency of any word in a corpus of text is inversely proportional to the rank of the word in the frequency table. This roughly holds for any text, for example it describes the distribution of the words in this paper.} to be the most prominent amongst them. It has been shown, however, that Zipfian power laws in linguistics arise from more fundamental distributions with rapid exponential decay \cite{WentianLi}. So there is hope for Gaussian and near-Gaussian distributions for adjectives and verbs to be derived from first principles. We will not pursue this direction here, and instead proceed to develop concrete permutation invariant Gaussian models which we compare with the data. 

It is worth noting that, while much work on applications of matrix theory to data in physics focuses on eigenvalue distributions, there are several reasons why this is not the ideal approach here. 
The matrices corresponding to adjectives or intransitive verbs created by linear regression are real and not necessarily symmetric ($M$ is not equal to its transpose); hence, their eigenvalues are not necessarily real.\footnote{Note however that hermitian matrices have been used in the past for word representation purposes. In \cite{calco2015}, for example, a compositional model of meaning inspired by categorical quantum mechanics is presented, where the meaning of words is given by density matrices. The model has yet to be experimentally verified in large scale tasks, so we do not deal with it here; for a small-scale preliminary evaluation on textual entailment, see \cite{balkir}.}
 One could contemplate a Jordan block decomposition,  where in addition to the fact that the eigenvalues can be complex, one has to keep in mind that additional information about the matrix is present in  the sizes of the Jordan block. More crucially, since our approach is guided by $S_D$ symmetry, we need to keep in mind that general base changes required to bring a matrix into Jordan normal form are not necessarily in $S_D$.  The natural  approach we are taking  consists in considering all possible $S_D$ invariant polynomial functions of the matrix $M$, and the averages of these functions constructed in a probability distribution  which is itself function of appropriate invariants. Specifically, in this paper we will consider the probability distribution to be a simple 5-parameter Gaussian, with a view to cubic and quartic perturbations thereof, and we will test the viability of this model by comparing to the data. 

Perturbed Gaussian matrix statistics can be viewed as the zero dimensional reduction of four dimensional quantum field theory, which 
is used to describe particle physics, e.g. the standard model. The approach to the data we describe in more detail in the next section is 
the zero dimensional analog of using effective quantum field theory to describe particle physics phenomena, where symmetry (in the 
present case $S_D$) plays an important role.

\section{The 5-parameter Gaussian model }\label{sec:5parameters} 

We consider a simple $S_D$ invariant Gaussian matrix model. The measure $dM$ is a standard measure on the $D^2$ matrix variables given in Section \ref{matrixmeasure}. This is multiplied by an exponential of a quadratic function of the matrices. The parameters $J^0, J^S$ are coefficients of  terms linear in the  diagonal and off-diagonal matrix elements respectively. The parameter $ \Lambda $ is the coefficient of  the square of the diagonal elements, while $ a , b $  are coefficients for off-diagonal elements. The partition function of the model is 
\bea\label{themodel} 
  \cZ ( \Lambda ,  a, b , J^0 , J^S  )&&  = \int dM e^{ - { \Lambda \over 2 } \sum_{ i = 1 }^D   M_{ii}^2  - { 1 \over 4 } ( a + b )  \sum_{ i < j } 
  ( M_{ij}^2  + M_{ji}^2 ) } \cr 
  && \hskip1.5cm e^{ -  { 1 \over 2 } ( a - b ) \sum_{ i < j } M_{ij} M_{ji}  +  J^0  \sum_{ i} M_{ii} + J^S  \sum_{ i <  j }  ( M_{ij} + M_{ ji } )  }  
\eea
The observables of the model are $ S_D $ invariant polynomials in the matrix variables:
\bea 
f ( M_{ i , j }  ) = f ( M_{ \sigma (i) , \sigma (j) } ) 
\eea

At quadratic order there are $11$ polynomials, which are listed in Section \ref{ElevenInvariants}. 
We have only used three of these invariants in the model above. The most general  matrix model compatible with  $ S_D$ symmetry 
 would consider all the eleven parameters and allow coefficients for each of them. In this paper, we restrict attention to  the simple 
5-parameter model, where the integral factorizes into $D $ integrals for the diagonal matrix elements and $ D ( D-1)/2$ integrals for the 
off-diagonal elements. Each integral for a diagonal element is a 1-variable integral. For each $( i  , j) $ with $ i < j$, we have an integral over 2 variables.

Expectation values of $f ( M ) $ are computed as 
\bea 
\langle f ( M ) \rangle \equiv { 1 \over \cZ } \int dM f ( M ) \text{EXP}
\eea
where EXP is the exponential term in (\ref{themodel}).  
In the following we give expressions for a set of linear, quadratic, cubic and quartic expectation values computed from theory. 
The computation follows standard techniques from  the path integral approach to quantum field theory. This 
involves introducing  sources $ J_{ij } $ for all the matrix elements and computing the general Gaussian integrals as function of
all these sources. The formula is given in equation (\ref{genfun}). Taking appropriate derivatives of the result gives the  expectation values of the observables. 

Since the theory is Gaussian, all the correlators can be given by Wick's theorem in terms of the linear and quadratic expectation values, as below:
\bea 
\langle M_{ij} \rangle  & = &  { 2 J^S  \over a }  ~~~ \hbox{ for all } ~~~  i \ne j \cr 
\langle M_{ii} \rangle  & = &  \Lambda^{-1} J^0 
\eea

For quadratic averages we have:
\bea 
\langle M_{ii} M_{jj} \rangle  & = &   \langle M_{ii} \rangle \langle M_{jj} \rangle ~ +~  \delta_{ij} \Lambda^{-1} \cr 
\langle M_{ij} M_{kl} \rangle  & = &  \langle M_{ij}\rangle \langle  M_{kl} \rangle + ( {  a^{-1} } + {  b^{-1}  } ) \delta_{ik} \delta_{jl} + ( {  a^{-1} } - {  b^{-1}  } ) \delta_{ il}  \delta_{ jk } 
\eea

From the above it also follows that:
\bea 
\langle M_{ii} M_{jj} \rangle_c ~ & = & ~ \delta_{ ij} \Lambda^{-1} \cr 
\langle M_{ij} M_{ ij}\rangle_c  ~ &=& ~  ( {  a^{-1}  } + {  b^{-1}  } )  ~~ \hbox{ for } ~~ i \ne j \cr
\langle M_{ij} M_{ji } \rangle_c ~ & = & ~  ( {  a^{-1}  } - {  b^{-1}  } ) ~~~ \hbox{ for  } ~~ i \ne j 
\eea

\subsection{Theoretical results for $ S_D$ invariant observables} 

In this section we give the results for expectation values of observables, which we will need for comparison to the experimental data, i.e. to averages over the collection of word matrices in the dataset described in Section \ref{sec:dataset}. The comparison of the linear and quadratic averages are used to fix the parameters $ J_0 ,J^S , \Lambda , a , b $. 
These parameters are then used to give the theoretical prediction for the higher order expectation values, which are compared with the experiment. 

\subsubsection{Linear order}

\bea\label{ModLin} 
&& M_{ d : 1} = \left\langle \sum_{ i } M_{ii} \right\rangle = \langle tr M \rangle = \Lambda^{-1} J^0 D \cr 
&& M_{o : 1} =  \left\langle \sum_{ i \ne  j  } M_{ ij } \right\rangle =  { 2 D ( D -1 ) \over a } J^S 
\eea

\subsubsection{Quadratic order}

\begin{align}\label{ModQuad} 
M_{ d : 2 } = \sum_{ i } \langle M_{ii}^2 \rangle & = \sum_i \langle M_{ii} \rangle \langle M_{ii} \rangle + \sum_i \Lambda^{-1} \cr 
                                     & = D \Lambda^{-2} ( J^0)^2  + D \Lambda^{-1}  \cr 
M_{o : 2,1 } = \sum_{ i \ne j } \langle M_{ij} M_{ ij} \rangle & = \sum_{ i \ne j } \langle M_{ij} \rangle \langle M_{ij } \rangle + \sum_{ i \ne j } ( a^{-1} + b^{-1} ) \cr 
                                                    &=  { D ( D-1)  }  \left ( {  4 (J^S)^2 a^{-2}  }  + ( a^{-1} + b^{-1}  )  \right ) \cr 
M_{ o : 2,2} = \sum_{ i \ne j } \langle M_{ij} M_{ji} \rangle &=  { D ( D-1) } \left ( 4 (J^S)^2 a^{-2} + ( a^{-1} - b^{-1} )  \right )  
\end{align} 

\subsubsection{Cubic order}

\begin{align}\label{ModCub} 
 M_{ d: 3} \equiv  \sum_i \langle M_{ii}^3 \rangle & =   \sum_{ i } \langle M_{ii} \rangle^3 + 3 \sum_{ i } \langle M_{ii}^2 \rangle_c  ~ \langle M_{ii} \rangle \cr 
 & =  D \Lambda^{-3} (  J^0 )^3 + 3 D \Lambda^{-2 }  ( J^0 ) \cr 
M_{ o : 3 ,  1} \equiv  \sum_{ i \ne j } \langle M_{ij}^3 \rangle & = \sum_{ i \ne j } \langle M_{ij}\rangle^3 + 3 \sum_{ i \ne j } \langle M_{ij} M_{ij} \rangle_c  \langle M_{ij} \rangle \cr 
  & =  D ( D-1) \left (    ({ 2 J_s \over a })^3 + { 6 J_s \over a } ( a^{-1} + b^{-1} ) \right )  \cr    
 M_{o: 3,  2 } = \sum_{ i \ne j \ne k } \langle M_{ij} M_{jk} M_{ki } \rangle 
  & =  8 D ( D-1 ) ( D-2 )   ( J^S)^3 a^{-3}  
\end{align} 

\subsubsection{Quartic order}

\begin{align}\label{ModQuart} 
& M_{ d: 4} = \sum_i \langle M_{ii}^4 \rangle  =  D \Lambda^{-2} \bigg ( \Lambda^{-2} ( J^0 )^4 + 6 \Lambda^{-1} ( J^0)^2 + 3 \bigg )   \cr 
&  M_{ o ;  4,1 }  = \sum_{ i \ne j }  \langle M_{ij }^4 \rangle =  \sum_{ i \ne j } \langle M_{ ij} \rangle^4 + 6 \langle M_{ij} M_{ ij} \rangle_c \langle M_{ij}\rangle^2 + 3 \langle M_{ij} M_{ ij} \rangle_c^2 \cr 
& = { D ( D-1) } \left ( \left ( {  2J_s \over a } \right )^4 + 6 \left ( { 2 J_s \over a } \right )^2  ( a^{-1} + b^{-1} ) +     3 ( a^{-1} + b^{-1} )^2 
\right )  \cr 
 &  M_{o: 4, 2} = \sum_{ i \ne j \ne k \ne l }   \langle M_{ij} M_{jk} M_{kl} M_{li} \rangle   = 16 D ( D-1) ( D-2 ) ( D-3 ) a^{-4} ( J^S )^4 \cr 
\end{align}

\section{Comparison of Gaussian models and linguistic data} 
\label{sec:comparison}

An ideal theoretical model would be defined by a 
partition function 
\bea 
\cZ( M ) = \int d M  e^{-  S (  M ) } 
\eea
with some appropriate function  $ S ( M ) $ (the Euclidean action of a zero dimensional matrix quantum field theory)  
such that theoretical averages 
\bea 
 \langle f ( M ) \rangle  = { 1 \over \cZ }  \int d M  e^{-  S (  M ) }  f ( M ) 
\eea
would agree with experimental averages 
\bea 
 \langle f ( M ) \rangle_{EXPT} = {  1 \over \rm { Number ~ of ~  words}  } \sum_{ words } f_{word}  (M)
\eea
to within the intrinsic uncertainties in  the data, due to limitations such as the small size of the dataset. 

In the present investigation we are comparing a Gaussian theory with the data, which is well motivated by the plots shown earlier in Figure \ref{fig:GME}. The differences between theory and experiment can be used to correct the Gaussian theory, by adding cubic and quartic terms (possibly higher) to get better approximations to the data. This is in line with how physicists approach elementary particle physics, where the zero dimensional matrix integrals are replaced by higher dimensional  path integrals involving matrices, the quadratic terms in the action encode the particle content of the theory, and higher order terms encode small interactions in perturbative quantum field theory: a framework which works well for the standard model of particle physics.  

We use the data for the averages $M_{d:1}$, $M_{ o:1 }$ , $M_{d:2}$ , $M_{o:2,1}$, and $M_{o:2,2}$ 
to determine the parameters $ J_0 , \Lambda, J_s , a, b$  of the  Gaussian Matrix 
model for a range of values of $D$ (the number of context words)  ranging from $D=300$ to $D=2000$, increasing in steps of $100$.  Working with the adjective part of the dataset, we find:

\begin{align} 
{ J_0  \over D }  &=  1.31 \times 10^{-2}  \cr 
{ \Lambda \over D^2}  &=  2.86 \times 10^{-3} \cr 
{ J_s \over D } & =  4.51 \times 10^{-4}  \cr 
{ a \over D^2 } &= 1.95 \times 10^{ -3}  \cr 
{ b \over D^2 } & = 2.01 \times 10^{-3} 
\end{align}

\begin{figure}
  \centering
  \includegraphics[width=8cm]{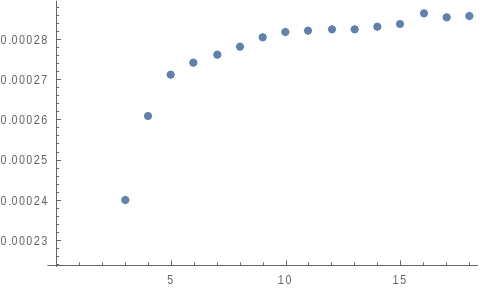}
  \caption{The ratio $ {\Lambda\over D^2} $ stabilzing at large $D$. }
  \label{fig:LamOvDsq}
\end{figure}

The plot in Figure \ref{fig:LamOvDsq} shows that the ratio $ { \Lambda \over D^2} $
approaches a constant as $D $ increases towards large values. Plots for 
the other ratios above show a similar stabilization. The calculation of 
the averages were repeated by permuting the set of 2000 context words, and repeating the calculation  for different  values of $D$. In these two experiments, the values of the parameters at an intermediate $D$ around $ 1200$ are compared. The two sets 
of context words only have a partial overlap, due to the fact that both come from the same $2000$ contexts. We find differences in the parameters of order one percent. We thus estimate that the random choice of context words results in an uncertainty 
of this order. 

Using these Gaussian parameters, we can calculate the expectation values of a number of 
cubic and quartic invariants in  the matrix model, which we then compare with the experimental values.  The difference between diagonal cubic correlators in theory and experiment is small. We have:
\bea 
  ( M_{d:3}^{\rm{ THRY } } /  M_{d:3}^{ \rm{EXPT} } ) = 0.57 
  \eea
indicating a percentage difference of $43 \% $. As a next step 
in the theory/experiment comparison, we would contemplate adding 
cubic terms to the Gaussian exponentials, following the philosophy 
of perturbative quantum field theory, adapted here to matrix statistics. 
We would then use these peturbations to obtain better estimates of cubic and higher order averages. The difference of $0.43$ can be used as a sensible estimate of 
the size of the perturbation parameter. Calculations of up to fifth order 
would then reach accuracies of around one percent, comparable to the one percent uncertainties discussed above. This is a reasonable order of perturbative calculation 
comparable to what can be  achieved in perturbative quantum field theory. The latter involves substantial additional complexity due to integrals over four-dimensional space-time momenta.  

Prior to engaging in detailed perturbation theory calculations, 
it is important to test the stabilization of the parameters as $D$ increases
 above 2000. Furthermore, in these experiments we have worked with a small sample of 273 adjectives. This should also be increased in order to test ensure that we are in a region of sufficiently large numbers, where universal features are likely to be manifest. 

For the quartic diagonal average we have:
\bea 
  ( M_{d:4}^{\rm{ THRY } } /  M_{d:4}^{ \rm{EXPT} } ) = 0.33 
\eea
with a percentage difference of $0.67$. While the data is again not Gaussian at the level of reliability, this is still a very realistic set-up for perturbation theory around the 
 Gaussian model.

For the simplest off-diagonal moments, the difference between experiment and theory is larger, but still within the realm of perturbation  theory:
\bea 
&&   ( M_{o:3,1}^{\rm{ THRY } } /  M_{o:3,1}^{ \rm{EXPT} } ) = 0.32 \cr 
&&  ( M_{o:4,1}^{\rm{ THRY } } /  M_{o:4,1}^{ \rm{EXPT} } ) = 0.47 
\eea
However, once we move to the more complex off-diagonal moments involving 
triple sums, the differences between theory and experiment start to become very substantial:
\bea
&&   ( M_{o:3,2}^{\rm{ THRY } } /  M_{o:3,2}^{ \rm{EXPT} } ) = 0.013\cr 
&&  ( M_{o:4,2}^{\rm{ THRY } } /  M_{o:4,2}^{ \rm{EXPT} } ) = 0.0084
\eea
In the framework of permutation symmetric Gaussian models we are advocating, this is in fact not surprising. As already mentioned, the 5-parameter Gaussian Matrix model we have  considered is  not the most general allowed by the symmetries. 
There are other quadratic terms we can insert into the exponent, for example:
\bea 
e^{ - c \sum_{ i \ne j \ne k  } M_{ ij} M_{ jk} } 
\eea
for some constant. 
This will lead to non-zero two-point averages:
\bea 
\langle M_{ ij} M_{ jk} \rangle - \langle M_{ij} \rangle \langle M_{jk} \rangle
\eea
By considering $c$ as a perturbation around the 5-parameter model in a limit of small $c$, we see that this will affect the theoretical calculation for 
\bea 
\sum_{ i\ne j \ne k } \langle M_{ij} M_{ jk} M_{ki} \rangle
\eea

A similar discussion holds for the matrix statistics for the verb part of the dataset. The parameters of the Gaussian model are now:

\bea 
{ J_0  \over D }  &=  1.16 \times 10^{-3} \cr 
{ \Lambda \over D^2}  &=  2.42 \times 10^{-3} \cr 
{ J_s \over D } & =  3.19  \times 10^{-4}  \cr 
{ a \over D^2 } &= 1.58  \times 10^{ -3}  \cr 
{ b \over D^2 } & = 1.62 \times 10^{-3} 
\eea

The cubic and quartic averages involving two sums over $D$ show departures from 
Gaussianity which are broadly within reach of a realistic peturbation theory approach:
\bea 
 && ( M_{d:3}^{\rm{ THRY } } /  M_{d:3}^{ \rm{EXPT} } ) = 0.54 \cr 
 &&  ( M_{d:4}^{\rm{ THRY } } /  M_{d:4}^{ \rm{EXPT} } ) = 0.30 \cr 
&&   ( M_{o:3,1}^{\rm{ THRY } } /  M_{o:3,1}^{ \rm{EXPT} } ) = 0.25 \cr 
&&  ( M_{o:4,1}^{\rm{ THRY } } /  M_{o:4,1}^{ \rm{EXPT} } ) = 0.48 
\eea 

The more complex cubic and quartic averages show much more siginificant differences between experiment and
theory, which indicates that a more general Gaussian should be the starting point of perturbation theory:
\bea 
&&   ( M_{o:3,2}^{\rm{ THRY } } /  M_{o:3,2}^{ \rm{EXPT} } ) = 0.010\cr 
&&  ( M_{o:4,2}^{\rm{ THRY } } /  M_{o:4,2}^{ \rm{EXPT} } ) = 0.006 
\eea

The most general quadratic terms compatible with invariance 
under $ S_D$ symmetry are listed in the Appendix \ref{sec:SDMatrixInvts}. While there are eleven of them, only  three were included
(along with the two linear terms) in the 5-parameter model. Taking into account some of the additional quadratic terms 
 in the exponential of (\ref{themodel})  will require a more complex theretical calculation in order  to arrive at the 
 predictions of the theory. While the 5-parameter integral can be factored into a product of integrals for each diagonal 
 matrix element and a product over pairs $ \{ ( i,j)  : i < j \}  $,  this is no longer the case with the more general Gaussian models. 
 It will require the diagonalization of a more complex bilinear form coupling the $ D^2$ variables $ M_{ij}$.  

In the Appendix  \ref{sec:SDMatrixInvts} we also discuss higher order invariants of general degree $k$ using representation theory of $ S_D$  and observe that these invariants are in correspondence with directed graphs. From a data-analysis perspective, the averages over a collection of word matrices of these invariants form the complete set of characteristics of the specific dataset. From the point of view of matrix theory, the goal is to find an appropriate weight of the form ``Gaussian plus peturbations'' which will provide an agreement with  all the observable averages to within uncertainties intrinsic to the data.   

In the ideal case, Gaussian models with low order perturbations would reproduce arbitrarily high order moments. 
In theoretical  particle physics, in many cases the  quantum fields are matrices, e.g. the gluons mediating the strong force
and quantum field theory involves doing  integrals over these fields which are parametrized by four space-time coordinates.  
The dimensional reduction of the quantum field theory to zero dimension gives a matrix theory. The fact that the standard model of particle 
physics is renormalizable means that the reduced matrix statistics of gluons and other particles involves only low order perturbations of Gaussian terms. It would be fascinating if language displays analogs of 
this renormalizability property. 

\section{Discussion and future directions} 
 
We find evidence that perturbed Gaussian models based on permutation invariants provide a viable approach to analyzing matrix data in tensor-based models of meaning. Our approach has been informed by matrix theory and analogies to particle physics. The broader lesson is that viewing language as a physical system and characterizing the universality classes of the statistics in compositional distributional models can provide valuable insights. In this work we analyzed the matrix data of words in terms of permutation symmetric Gaussian Matrix models. In such models, the continuous symmetries $ SO(D)$, $Sp(D)$, $U(D)$ typically encountered in physical systems involving matrices of size $D$, have been replaced by the symmetric groups $S_D$. The simplest 5-parameter Gaussian models compatible with this symmetry  were fitted to 5 averages of linear and quadratic $S_D$ invariants constructed from the word matrices. The resulting model was used to predict averages of a number of cubic and quartic invariants. Some of these averages were well within the realm of perturbation theory around Gaussians. However, others showed significant departures which motivates a more general study of Gaussian models and their comparison with linguistic data for the future. The present investigations have established a framework for this study which makes possible a number of interesting theoretical as well as data-analysis projects suggested by this work.   

An important contribution of this work is that it allows the characterization of any text corpus or word class in terms of thirteen Gaussian parameters: the averages of two linear and eleven quadratic matrix invariants listed in Appendix \ref{sec:SDMatrixInvts}. This can potentially provide a useful tool that facilitates research on comparing and analyzing the differences between:

\begin{itemize}
  \item natural languages (e.g. English versus French)
  \item literature genres (e.g. The Bible versus The Coran; Dostoyevski versus Poe; science fiction versus horror)
  \item classes of words (e.g. verbs versus adjectives)
\end{itemize}

Another interesting question that naturally arises in the light of this work is how the present theory can be used towards an improved tensor model of natural language semantics. The theory provides a notion of degrees of departure from Gaussianity, that could be potentially exploited in constructing the word matrices and tensors in order to address data sparsity problems and lead to more robust distributional representations of meaning.

Furthermore, while we have focused on matrices, in general higher tensors are also involved (for example, a ditransitive verb\footnote{A verb that takes two objects, one direct and one indirect, as the verb ``gave'' in ``I gave the policeman a flower''.} is a tensor of order 4). The theory of permutation invariant Gaussian matrix models can be extended to such tensors as well. For the  case of continuous symmetry, the generalization to tensors has been fruitfully studied \cite{GR2011} and continues to be an active subject of research in mathematics. Algebraic techniques for the enumeration and computation of corelators of tensor invariants \cite{tenscount} in the continuous symmetry models should continue to be applicable to $S_D$ invariant systems.  These techniques rely on discrete dual symmetries, e.g. when the problem has manifest  unitary group symmetries acting on the indices of one-matrix or multi-matrix systems, permutation symmetries arising from Schur-Weyl duality play a role. This is reviewed in the context of the AdS/CFT duality in \cite{SWrev}. When the manifest 
symmetry is $S_D$, the symmetries arising from Schur-Weyl duality will involve  partition algebras \cite{Martin96,HalvRam04}. 

As a last note, we would like to emphasize that while this paper draws insights from physics for analysing natural language, this analogy can also work the other way around. Matrix models are dimensional reductions of higher dimensional quantum field theories, describing elementary particle physics, which contain matrix quantum fields. In many cases, these models capture important features of the QFTs: e.g. in free conformal limits of quantum field theories, they capture the 2- and 3-point functions. An active area of research in theoretical physics seeks to explore the information theoretic content of quantum field theories \cite{CC04,NRT09}. It is reasonable to expect that the application of the common mathematical framework of matrix theories to language and particle physics will suggest many  interesting analogies, for example, potentially leading to new ways to explore complexity in QFTs by developing analogs of linguistic complexity.

 \bigskip 
 
 \begin{centerline} 
{\bf Acknowledgments} 
\end{centerline} 
\vskip.4cm 

The research of SR is supported by the  STFC Grant ST/P000754/1, String Theory, Gauge Theory, and Duality; and by a Visiting Professorship at the University of  the Witwatersrand, funded by a Simons Foundation grant awarded to the Mandelstam Institute for Theoretical Physics. DK and MS gratefully acknowledge support by AFOSR International
Scientific Collaboration Grant FA9550-14-1-0079. MS is also supported by EPSRC for Career Acceleration Fellowship EP/J002607/1.
We are grateful for discussions to  Andreas Brandhuber, Robert de Mello Koch, Yang Hao,  Aurelie Herbelot, Hally Ingram, Andrew Lewis-Pye, Shahn Majid and Gabriele Travaglini.   

\begin{appendix} 

\section{Gaussian Matrix Integrals: 5-parameter model}\label{App:MatrixIntegrals}

$M$ is a real $ D \times D $ matrix. $S$ and $A$ are the symmetric and anti-symmetric parts. 
\bea 
S = { M + M^{T} \over 2 } \cr 
A = { M - M^T \over 2 } 
\eea
Equivalently,
\bea 
&& S_{ ij } = { 1\over 2   } ( M_{ij} + M_{ji} ) \cr 
&&A_{ij  } = { 1 \over 2 } ( M_{ij}  - M_{ji} ) 
\eea
We have:
\bea 
&& S^T = S \cr 
&& A^T = - A \cr 
&& M = S + A 
\eea
The independent elements of $S$ are $ S_{ij} $ for $ i \le j$, i.e the elements along the diagonal $ S_{ii} $ and the 
elements above $ S_{ ij} $ for $ i  < j $. The independent elements of $A$ are  $ A_{ij} $ for $ i < j $. The diagonal elements are zero. 
Define:
\bea\label{matrixmeasure} 
dM = \prod_{ i =1}^D dS_{ii} \prod_{ i < j } dS_{ij} dA_{ij} 
\eea
We consider the Gaussian partition function 
\bea  Z ( \Lambda ,  B ; J  ) = \int dM e^{ - \sum_{ i } { \Lambda_i\over 2 }  M_{ii} - { 1 \over 2 } \sum_{ i < j }  (S_{ij} ,  A_{ij} ) B_{ij } ( S_{ij} , A_{ij})^T  }  ~~~  e^{ \sum_{ i} J_{ii} M_{ii} + \sum_{ i \ne j } J_{ij} M_{ij} }  
\eea
Here $ B_{ij}$ is a two by two matrix with positive determinant:
\bea
B_{ij} = \begin{pmatrix} a_{ij} & c_{ij} \\ c_{ij} & b_{ij} \end{pmatrix}  ~~~~~ \det ( B_{ij} ) = a_{ij} b_{ij} - c_{ij}^2  >0 
\eea
It defines the quadratic terms involving $ ( A_{ij } , B_{ij} ) $.
\bea 
(S_{ij} ,  A_{ij} ) B_{ij } ( S_{ij} , A_{ij} ) ^T   = a_{ij} S_{ij}^2 + b_{ij } A_{ij}^2  + 2 c_{ij} S_{ij} A_{ij} 
\eea
The condition $ \det B_{ij} > 0 $ ensures that the integral converges. Choosing these quadratic parameters of the Gaussian to be constants  $ c_{ij} = c,  a_{ij} = a , b_{ij} = b $ ensures that the model is permutation symmetric. For simplicity, we will also choose $ c=0$. 
The  linear terms (also called source terms)  can be re-written as:
\bea 
e^{ \sum_{ i \ne j } J_{ij} M_{ij} + \sum_i J_i M_{ii} } 
= e^{ \sum_{ i} J_{ii} M_{ii} + \sum_{ i < j } ( 2 J_{ij}^S S_{ij} + 2 J_{ij}^A A_{ij} ) }
\eea
where $ J_{ij}^S , J_{ij}^A $ are the symmetric and anti-symmetric parts of the source matrix. 
\bea
J_{ij}^S ={ 1 \over 2 } (  J_{ij } + J_{ji} ) \cr 
J_{ij}^A = { 1 \over 2 } ( J_{ij} - J_{ji } )  
\eea
Using a standard formula for multi-variable Gaussian integrals (see for example \cite{WikiGaussians}):
\bea\label{genfun}  
&& \cZ ( \Lambda  , B ; J )  = \sqrt{ ( 2 \pi)^{ N^2} \over \prod_i \Lambda_i \prod_{i < j } \det B_{ij} }
e^{ { 1\over 2} \sum_{ i } J_{ii} \Lambda_i^{-1} J_{ii} +\sum_{ i < j }  { 2 \over \det B_{ij } }  \left ( b_{ij } (J_{ij}^S )^2 + a_{ij} ( J_{ij}^A )^2  - 2 c_{ij }  J_{ij}^A J_{ij}^S \right ) } \cr 
&& 
\eea
For any function of the matrices $ f ( M ) $ the expectation value is defined by:
\bea 
\langle f (  M ) \rangle = { 1 \over \cZ } \int  dM ~ f ( M ) ~ \text{EXP} 
\eea
where EXP is the product of exponentials defining the Gaussian measure. 
Following standard techniques from the path integral approach to  quantum field theory, the expectation values are calculated using 
derivatives with respect to sources (see e.g. \cite{Peskin}).

\section{Counting $S_D$ invariant matrix polynomials}\label{sec:SDMatrixInvts} 

There are $11$ quadratic invariants in $M_{ij}$ which are invariant under $S_D$ ( $D \ge 4$). 
\bea\label{ElevenInvariants} 
&& \sum_{ i  } M_{ii}^2 \cr 
&& \sum_{ i \ne j  } M_{ ij}^2 ~~~ , ~~~ \sum_{ i \ne j } M_{ ij} M_{ji} \cr 
&& \sum_{ i \ne j } M_{ ii} M_{jj}  ~~~ , ~~~ \sum_{ i \ne j } M_{ii} M_{ ij} ~~ , ~~  \sum_{ i \ne j }  M_{ ij}M_{jj}  \cr 
&& \sum_{ i \ne j \ne k } M_{ ij }M_{ jk} , \sum_{ i \ne j \ne k } M_{ ij} M_{ ik} , \sum_{ i \ne j \ne k } M_{ij} M_{ kj } , \sum_{ i\ne j \ne k } M_{ ij} M_{kk}  \cr 
 && \sum_{ i \ne j \ne k \ne l } M_{ ij } M_{ kl }    
\eea

\begin{figure}[t!]

  \centering
 
  \begin{tabular}{cccccccc}
    
    ~\includegraphics[scale=0.50]{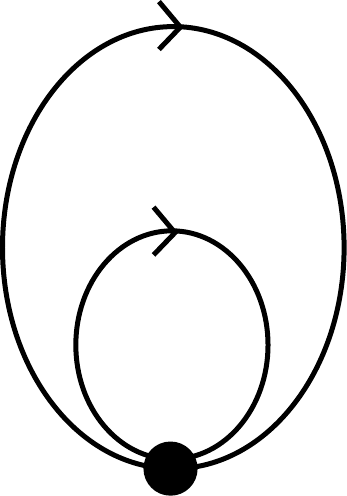}~ &
    ~\includegraphics[scale=0.50]{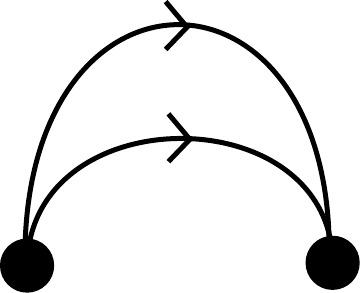}~ &
    ~\includegraphics[scale=0.50]{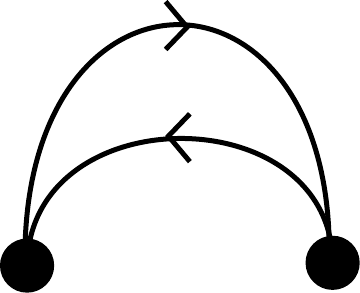}~ &
    ~\includegraphics[scale=0.50]{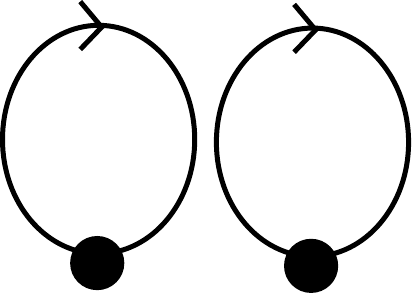}~ &    
    ~\includegraphics[scale=0.50]{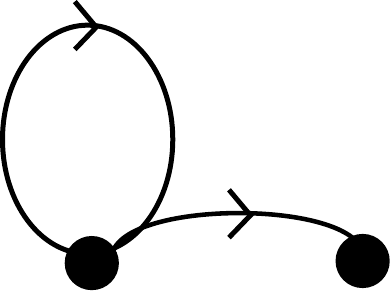}~ &    
    ~\includegraphics[scale=0.50]{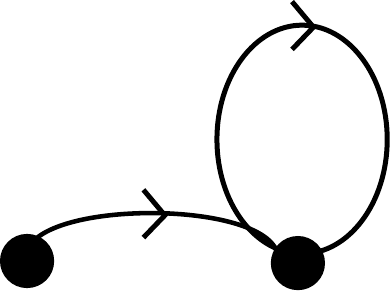}~  \\
    
    $\sum\limits_i M^2_{ii}$ & 
    $\sum\limits_{i\neq j} M^2_{ij}$ &
    $\sum\limits_{i\neq j} M_{ij}M_{ji}$ &
    $\sum\limits_{i\neq j} M_{ii}M_{jj}$ &
    $\sum\limits_{i\neq j} M_{ii}M_{ij}$ &
    $\sum\limits_{i\neq j} M_{ij}M_{jj}$ 
  
  \end{tabular}  

  \begin{tabular}{cccc}    
    ~\includegraphics[scale=0.50]{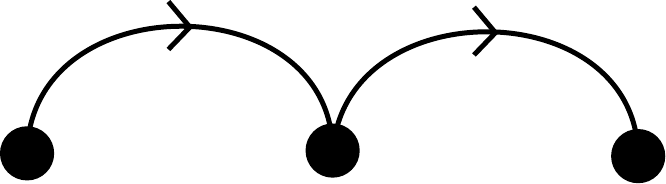}~ &
    ~\includegraphics[scale=0.50]{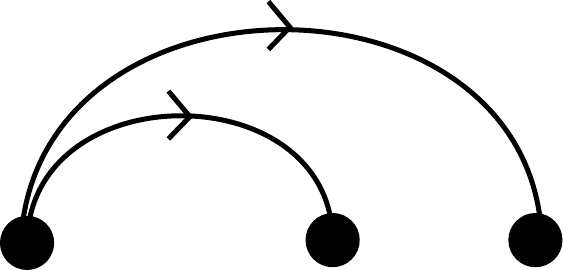}~ &
    ~\includegraphics[scale=0.50]{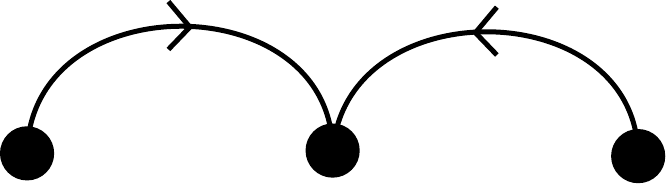}~ &
    ~\includegraphics[scale=0.50]{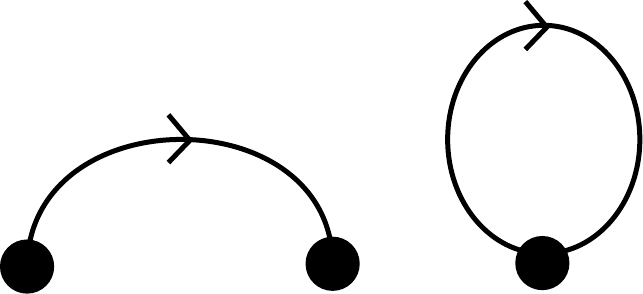} \\
    
    $\sum\limits_{i\neq j \neq k} M_{ij}M_{jk}M_{ki}$ &
    $\sum\limits_{i\neq j \neq k} M_{ij}M_{ik}$ &
    $\sum\limits_{i\neq j \neq k} M_{ij}M_{kj}$ &
    $\sum\limits_{i\neq j \neq k} M_{ij}M_{kk}$
  \end{tabular}
  
  \begin{tabular}{c}
    ~\includegraphics[scale=0.50]{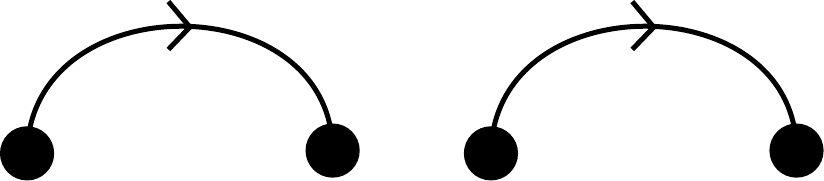} \\
    $\sum\limits_{i\neq j \neq k \neq l} M_{ij}M_{kl}$
  \end{tabular}

  \caption{$S_D$ invariant functions and graphs illustrated for quadratic invariants}
  \label{fig:Invts-Graphs}
\end{figure}

The sums run over $ 1 \dots D$. For $D=3$, the last invariant is zero. For $D=2$, there are $6$ invariants. In general we are interested in 
$D$ being large. By associating $M_{ij}$ to a directed edge connecting vertex $i$ to vertex $j$, the above list corresponds to counting 
graphs. This connection between directed graphs and invariants is illustrated in Figure  \ref{fig:Invts-Graphs}. 

There is a representation theoretic way to obtain the counting formula as a function of the degree $k$ of invariants (equal to $2$ above) 
and the dimension $D$. In our simple Gaussian theoretical model we have two linear terms along with the $3$ quadratic terms
in the first two lines of (\ref{ElevenInvariants}).   A general Gaussian theory compatible with $S_D$ symmetry would 
take into account all the invariants.  There are two linear invariants which can be averaged over the words. 
 The experimental input into the Gaussian model would consist of the averages 
for all the invariants. Thus permutation symmetry leads to  the characterization of matrices in tensor-based models of meaning
by means of 13 Gaussian parameters. From a purely experimental point of view, it is interesting to also characterize the matrix data 
using further higher order invariants. Below, we explain the representation theory approach to the counting of higher order invariants. 

Let $V_D$ be the $D$-dimensional permutation representation, also called  the natural  representation, of
$S_D$. The counting of invariants of degree $k$ is the same as counting of 1-dimensional representations of 
$S_D$ in the decomposition into irreducibles of 
\bea 
 Sym^k (  V_D \otimes V_D ) 
\eea
This can be expressed in terms of characters. Define:
\bea 
V_{D; 2}  = V_D \otimes V_D 
\eea
Given the linear  for $ \sigma $ in $ V_D$ which we denote as $ \cL_{D}  ( \sigma )$, 
the linear operator  in $ V_{D;2}  $ is 
\bea 
\cL_{D; 2 } ( \sigma ) = \cL_{ D}  ( \sigma )  \otimes \cL_{ D}  ( \sigma )  
\eea
The tensor product $ V_{D;2}^{ \otimes k } $ has an action of $ \sigma $ as 
\bea 
\cL_{ D;2; k } ( \sigma ) = \cL_{D; 2 } ( \sigma )  \otimes \cdots \otimes \cL_{ D;2 } ( \sigma ) 
\eea
where we are taking $k$ factors. 
The symmetric subspace of $ V_{ D ;  2}^{ \otimes k } $ is obtained by an action of permutations $ \tau \in S_k $, 
which involves permutating the $k$ tensor factors. 
The dimension of this subspace is:
\bea
Dim ( D , k ) && = { 1 \over k! D! } \sum_{ \sigma \in S_D} \sum_{ \tau \in S_k  } tr_{ V_{ D;2}^{\otimes k} } (  \cL_{ D;2; k } ( \sigma ) \tau ) \cr 
  && = { 1 \over D! k!  } \sum_{ \sigma \in S_D } \sum_{ \tau \in S_k } \prod_{ i=1 }^k  ( tr_{ V_{D;2}  } ( \sigma^i ) )^{ C_i ( \tau ) }    
\eea
Now use the fact that:
\bea 
tr_{ V_{ D ; 2 } } (  \cL_{D; 2 } ( \sigma )  ) =  ( tr_{ V_D } ( \cL_{ D } ( \sigma ) )^2  = ( C_1 ( \sigma )  )^2
\eea
The last step is based on the observation that the trace of a permutation in the natural representation is equal to the number of one-cycles in the permutation. We also need:
\bea 
 tr_{ V_D } ( \cL_{ D } ( \sigma^i  ) )  = \sum_{ l |i }  l C_l ( \sigma )  
\eea
This is a sum over divisors of $i$. 
We conclude:
\bea 
Dim ( D , k ) && = { 1 \over D ! k!  } \sum_{ \sigma \in S_D } \sum_{ \tau \in S_k } \prod_{i=1}^k    ( \sum_{ l|i }  l C_l ( \sigma ) )^{ 2  C_i ( \tau ) } 
\eea
The expression above is a function of the conjugacy classes of the permutations $ \sigma , \tau $. These conjugacy classes are 
partitions of $ D , k $ respectively, which we will denote by $ p = \{ p_1 , p_2 , \cdots , p_D \}$ and $ q = \{ q_1 , q_2 , \cdots , q_D \} $
obeying $ \sum_{ i } i p_i = D  , \sum_{ i } i q_i = k$. 
 Thus 
\bea 
Dim ( D , k )  = { 1 \over D ! k! } \sum_{ p \vdash D } \sum_{ q \vdash k }   { D! \over \prod_{i=1}^D i^{ p_i} p_i! } { k! \over \prod_{ i=1 }^{ k}  i^{ q_i} q_i! } \prod_{i =1 }^k \left ( \sum_{ l | i }  l p_l \right )^{ 2 q_i}  
\eea
For fixed degree $k$  of  the invariants, as $D$ increases the number stabilizes once $D$ reaches $2k$. This is clear from the realization of these numbers in terms of counting of graphs or matrix invariants.  Hence the simplest formula 
for the number of invariants as a function of $k$ is 
\bea 
Dim ( 2k , k )   = \sum_{ p \vdash 2k  } \sum_{ q \vdash k }   { 1 \over \prod_{i=1}  i^{ p_i + q_i } p_i! q_i! }   \prod_{i =1 }^k \left ( \sum_{ l |i  }  l p_l \right )^{ 2 q_i}  
\eea
Doing this sum in Mathematica, we find that the number of invariant functions at $ k=2,3,4,5,6$ are $11, 52, 296 , 1724, 11060 $. 
These are recognized as the first few terms in the OEIS series A052171 which counts graphs (multi-graphs with loops on any number of nodes). 
The graph theory interpretatation follows by thinking about $ M_{ij} $ as an edge of a graph. 

The decomposition of $ V_D^{ \otimes k  } $, and the closely related problem $ V_H^{ \otimes k } $ where $V_H $ is the 
non-trivial irrep of dimension $ D-1$ in $V_D$, have been studied in recent mathematics literature \cite{BHH} and are related to Stirling numbers. 
Some aspects of these decomposition numbers were studied and applied to the construction of supersymmetric states in quantum field theory  \cite{BHR}. 

\section{Dataset}
\label{app:dataset}

Below we provide the list of the 273 adjectives and 171 verbs for which matrices were constructed by linear regression, as explained in Section \ref{sec:lr}.

\subsection{Adjectives}

\flushleft
1st, 2nd, actual, adequate, administrative, adult, advanced, African, agricultural, alternative, amazing, ancient, animal, attractive, audio, Australian, automatic, beautiful, biological, blue, brief, broad, Canadian, catholic, cell, cheap, chemical, chief, Chinese, Christian, civil, classic, classical, clinical, coastal, cold, competitive, complex, comprehensive, considerable, constant, contemporary, content, continuous, conventional, core, corporate, correct, creative, criminal, critical, cultural, daily, dark, dead, deep, detailed, digital, distinct, diverse, domestic, double, dramatic, dry, Dutch, dynamic, east, educational, electric, electrical, electronic, emotional, entire, environmental, equal, essential, exact, exciting, exclusive, existing, experienced, experimental, extensive, external, extra, fair, fantastic, fast, favourite, federal, fellow, female, fine, flat, foreign, formal, fourth, fresh, friendly, front, fundamental, game, genetic, global, Greek, green, ground, half, head, healthy, heavy, historic, historical, hot, huge, ideal, immediate, impressive, improved, increased, Indian, industrial, initial, inner, innovative, integrated, interactive, internal, Iraqi, Irish, Israeli, Italian, Japanese, Jewish, joint, key, lead, leading, level, library, light, limited, literary, live, London, lovely, mainstream, male, mass, massive, material, maximum, medieval, medium, mental, minimum, minor, minute, mixed, mobile, model, monthly, moral, multiple, musical, Muslim, name, narrow, native, near, nearby, negative, net, nice, north, northern, notable, nuclear, numerous, official, ongoing, operational, ordinary, organic, outdoor, outstanding, overall, overseas, part, patient, perfect, permanent, Polish, positive, potential, powerful, principal, prominent, proper, quality, quick, rapid, rare, reasonable, record, red, related, relative, religious, remote, residential, retail, rich, Roman, royal, rural, Russian, safe, scientific, Scottish, secondary, secret, selected, senior, separate, serious, severe, sexual, site, slow, soft, solid, sound, south, southern, soviet, Spanish, specialist, specified, spiritual, statutory, strange, strategic, structural, subsequent, substantial, sufficient, suitable, superb, sustainable, Swedish, technical, temporary, tiny, typical, unusual, upper, urban, usual, valuable, video, virtual, visual, website, weekly, welsh, west, western, Western, wild, wonderful, wooden, written

\subsection{Verbs}

accept, access, acquire, address, adopt, advise, affect, aim, announce, appoint, approach, arrange, assess, assist, attack, attempt, attend, attract, avoid, award, break, capture, catch, celebrate, challenge, check, claim, close, collect, combine, compare, comprise, concern, conduct, confirm, constitute, contact, control, cross, cut, declare, define, deliver, demonstrate, destroy, determine, discover, discuss, display, draw, drive, earn, eat, edit, employ, enable, encourage, enhance, enjoy, evaluate, examine, expand, experience, explain, explore, express, extend, face, facilitate, fail, fight, fill, finish, force, fund, gain, generate, grant, handle, highlight, hit, hope, host, implement, incorporate, indicate, influence, inform, install, intend, introduce, investigate, invite, issue, kill, launch, lay, limit, link, list, love, maintain, mark, match, measure, miss, monitor, note, obtain, organise, outline, own, permit, pick, plan, prefer, prepare, prevent, promote, propose, protect, prove, pull, purchase, pursue, recognise, recommend, record, reflect, refuse, regard, reject, remember, remove, replace, request, retain, reveal, review, save, secure, seek, select, share, sign, specify, state, stop, strengthen, study, submit, suffer, supply, surround, teach, tend, test, threaten, throw, train, treat, undergo, understand, undertake, update, view, walk, watch, wear, welcome, wish

\end{appendix} 

\bibliographystyle{unsrt}  
\bibliography{refs}

\end{document}